\documentclass[letterpaper]{article} 
\def\year{2022}\relax
\usepackage{aaai22}  
\usepackage{times}  
\usepackage{helvet}  
\usepackage{courier}  
\usepackage[hyphens]{url}  
\usepackage{graphicx} 
\usepackage{natbib}  
\usepackage{caption} 
\DeclareCaptionStyle{ruled}{labelfont=normalfont,labelsep=colon,strut=off} 
\usepackage{algorithm}
\usepackage{algorithmic}

\newcommand{\hush}[1]{}
\hush{DEMO: This text disappears}

\usepackage{amsmath}
\usepackage{colortbl}

\title{Data-Driven Mitigation of Adversarial Text Perturbation}

\nocopyright 

\author{
    Rasika Bhalerao\textsuperscript{\rm 1},
    Mohammad Al-Rubaie\textsuperscript{\rm 2},
    Anand Bhaskar\textsuperscript{\rm 2},
    Igor Markov\textsuperscript{\rm 2}
}
\affiliations{
    \textsuperscript{\rm 1}New York University,
    \textsuperscript{\rm 2}Facebook, Inc.\\
    rasikabh@nyu.edu,
    mti, anandb, imarkov@fb.com
}

\begin{document}
\maketitle
\begin{abstract}
Social networks have become an indispensable part of our lives, with billions of people producing ever-increasing amounts of text.
At such scales, content policies and their enforcement become paramount.
To automate moderation, questionable content is detected by Natural Language Processing (NLP) classifiers.
However, high-performance classifiers are hampered by misspellings and adversarial text perturbations.
In this paper, we classify intentional and unintentional adversarial text perturbation into ten types and propose a deobfuscation pipeline to make NLP models robust to such perturbations.
We propose Continuous Word2Vec (CW2V), our data-driven method to learn word embeddings that ensures that perturbations of words have embeddings similar to those of the original words.
We show that CW2V embeddings are generally more robust to text perturbations than embeddings based on character ngram\hush{: the cosine distance between the embeddings of a word and a version of the word perturbed by swapping adjacent characters is 0.175 that of a random pair of words with CW2V, while it is 0.351 that of a random pair of words with character ngram-based word embeddings}.
Our robust classification pipeline combines deobfuscation and classification, using proposed defense methods and word embeddings to classify whether Facebook posts are requesting engagement such as \textit{like}s.
Our pipeline results in engagement bait classification that goes from 0.70 to 0.67 AUC with adversarial text perturbation, while character ngram-based word embedding methods result in downstream classification that goes from 0.76 to 0.64.
\end{abstract}
\section{Introduction}

\begin{figure}[ht]
\includegraphics[width=0.45\textwidth]{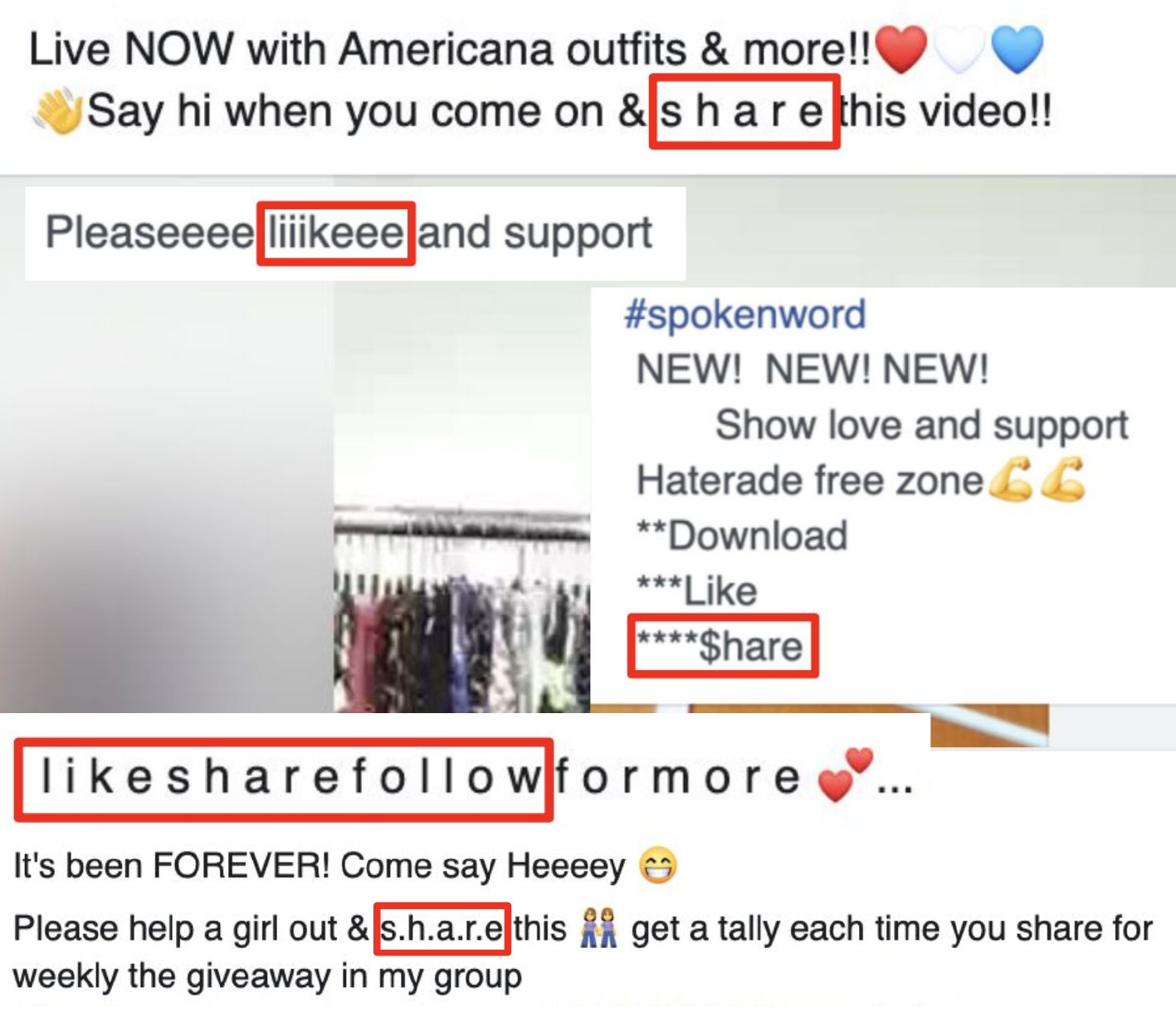}
\vspace{-3mm}
\caption{\label{figure:posts} Examples of Facebook posts adversarially perturbed to avoid detection and demotion for engagement bait.}
\vspace{-2mm}
\end{figure}

Social media hosts billions of active users, and enforcing
content policies is critical to user experience. Even with better-designed policies, such a scale requires automated moderation. Unfortunately, the Natural Language Processing (NLP) classifiers used for questionable content can be ($i$) confused by inadvertent typos, and ($ii$) deliberately manipulated by perturbing relevant text while preserving its human-readable appearance.

The motivation for our work can be illustrated by the following use case. Consider an ML classifier that detects Facebook posts that fall under \textit{engagement bait}, i.e., baiting the viewer to \textit{like}, \textit{comment}, etc. We investigate how this classifier can be fooled through adversarial text perturbations, and then we develop methods to mitigate such text perturbations.
Figure~\ref{figure:posts} shows illustrative examples of engagement bait Facebook posts that avoid detection by perturbing their text.
Adversarial text perturbations likely impact other classifiers, and our methods should be helpful more broadly than the immediate context used by our experiments.

In this paper, we identify ten common types of adversarial text perturbation based on our analysis of Facebook posts.
To address these perturbations, we develop rule-based string manipulations deobfuscation and trainable word embeddings such that words with perturbations have embeddings similar to those of the original unperturbed words.
The two rule-based string manipulations are (1) the Alternating Characters Defense (ACD), which detects and filters out alternating obfuscating characters (e.g. '-' in 't-e-x-t'), and (2) Unicode Canonicalization (UC), which uses a mapping to convert Unicode confusables to their ascii equivalents.
We assemble our defense methods into a deobfuscation pipeline and evaluate how well it detects harmful Facebook posts.

Our trainable word embeddings, Continuous Word2Vec (CW2V), strive to ensure that perturbed words have embeddings similar to their unperturbed versions. We accomplish this by treating words as continuous rather than discrete inputs, so a minor perturbation of the string also results in only a minor perturbation in the embedding.
We make the model originally proposed for Word2Vec \cite{mikolov2013efficient} more robust to perturbations by modifying the inputs and outputs to address our objective that words with similar spelling have similar embeddings.
Instead of passing words as one-hot encodings of the tokens in a vocabulary, we pass words as vectors that encode the string distances between the given word and each word in a specially selected index of words.
The model is then trained on these inputs the same way as in Word2Vec, resulting in embeddings that reflect both the spelling and the meaning of each word.

Cosine distances between our embeddings have a correlation of 0.77 with Levenshtein distances between words, confirming that small text changes only slightly change our embeddings. For embeddings trained using a prior character ngram-based method, the correlation is 0.05.
\hush{We show that our embeddings improve robustness of engagement bait classification to text perturbation. Using a standard character ngram-based word embedding method results in engagement bait classification with a 0.76 AUC and a 0.64 AUC with text perturbation, and our CW2V embedding method results in classification with a 0.70 AUC and a 0.67 AUC with text perturbation.}

This paper offers four key contributions:
\begin{enumerate}
    \item A taxonomy of text perturbations based on prior literature and additional ones that we discovered.
    \item A method called Continuous Word2Vec (CW2V) to construct word embeddings such that words with similar spelling or meaning map to similar vectors.
    \item A deobfuscation pipeline using CW2V and two other rule-based defense methods to improve the robustness of NLP classifiers against these perturbation types.
    \item Empirical impact measurements for perturbations and defenses in the classification pipeline, with comparison to prior state-of-the-art methods.
\end{enumerate}

\begin{table*}[ht]
\renewcommand{\arraystretch}{1.5}
\begin{center}
\begin{tabular}{|>{\columncolor[gray]{0.8}}p{4.5cm}p{1.4cm}>{\columncolor[gray]{0.8}}p{4.1cm}p{5.0cm}|}
\hline
\sc 
Perturbation type & \sc Defense & \sc Example & \sc Definition \\
\hline
Combined Unicode & ACD & P.l.e.a.s.e l.i.k.e a.n.d s.h.a.r.e & Insert a Unicode character between each original character. \hush{Generalized version of [Zero-width] space separation below.} \\
Fake punctuation & CW2V & Pleas.e lik,e and shar!e & Randomly add zero or more punctuation marks between characters. \\
Neighboring key & CW2V & Plwase lime and sharr & Replace characters with keyboard-adjacent characters. \\
Random spaces & CW2V & Pl ease lik e and sha re & Randomly insert zero or more spaces between characters. \\
Replace Unicode & UC & Pleãse lîke and sharê & Replace characters with Unicode look-alikes. \\
Space separation & ACD & Please l i k e and s h a r e & Place spaces between characters. \\
Tandem character obfuscation & UC & \small PLE/\textbackslash SE LIKE /\textbackslash ND SH/\textbackslash RE & Replace individual characters with multiple characters that together look like the original. \\
Transposition & CW2V & Plaese like adn sahre & Swap adjacent characters. \\
Vowel repetition and deletion & CW2V & Pls likee nd sharee & Repeat or delete vowels. \\
Zero-width space separation & ACD & Please like and share & Place zero-width spaces (Unicode character 200c) between characters. \\
\hline
\end{tabular}
\end{center}
\caption{Perturbations performed on the string ``Please like and share''. CW2V is the word embedding method proposed in this paper. ACD (Alternating Characters Defense) is a string manipulation to detect when the characters of a word are all separated by a repeating character (e.g. '-' in 't-e-x-t') and filter out that character. UC (Unicode Canonicalization) uses a mapping to convert Unicode confusables to their ascii equivalents.}
\label{table:perturbation_types}
\renewcommand{\arraystretch}{1}
\end{table*}

\section{Background and Problem Analysis}

\paragraph{Ethics}
We preface our methods with the warning that the context, ethics, and harms of the content detection algorithm itself must be carefully considered.
With this research, we do not aim to empower governments or law enforcement to target vulnerable groups.
We acknowledge that certain populations may be using text obfuscation and perturbation to protect themselves from surveillance, de-platforming, or detection in repressive regimes.
We urge those employing mitigation for adversarial text perturbation in their classification pipelines to thoroughly investigate the ethics of their applications before deployment.

\paragraph{Prior work on text perturbations} distinguishes several types of text perturbation, quantifies their effect on downstream classifiers, and publishes tools for simple yet effective adversarial text manipulation \cite{vijayaraghavan,eger-etal-2019-text,8424632,Li_2019}. Prior work also shows that simply training and testing on adversarially perturbed data does not improve downstream performance against these attacks because of the combinatorial explosion of possible obfuscations \cite{belinkov2017synthetic}.

\paragraph{Prior approaches to mitigating adversarial text perturbation} include inferring various word vectors from the context of each out-of-vocabulary word and finding a word in the dictionary that minimizes (1) the distance between its string and the string of the out-of-vocabulary word in question and (2) the distance between its vector and the context-inferred word vector \cite{zhou-etal-2019-learning,10.5120/ijca2019919384,fivez-etal-2017-unsupervised}. Methods for calculating word vectors and string distances vary, and the way in which the two objectives were optimized together vary. \citet{roben} cluster words in a corpus such that each cluster contains possible perturbed versions of the same word, and then use a single token to represent the words in each cluster. While their method fixes the input text so that the downstream model does not need to be robust to perturbations, our word embedding method takes context into account when learning embeddings, and our deobfuscation does not depend on having seen the right word somewhere in the training set. Choosing the best word by spelling without context is difficult when the test set contains words unseen in training; for example, there are 26 words in the English dictionary that are a single character away from the word \textit{like}\footnote{\textit{alike}, \textit{bike}, \textit{Dike}, \textit{fike}, \textit{glike}, \textit{Hike}, \textit{yike}, \textit{kike}, \textit{leke}, \textit{lice}, \textit{lige}, \textit{like}, \textit{liked}, \textit{liken}, \textit{liker}, \textit{likes}, \textit{lile}, \textit{lime}, \textit{lire}, \textit{lite}, \textit{live}, \textit{loke}, \textit{Mike}, \textit{Nike}, \textit{Pike}, \textit{sike}, \textit{tike}}, and while defining the mapping of perturbed words to correct words during training helps, it requires that words in the test set outside the mapping defined during training are not in the mapping.

\paragraph{Related work on word embeddings} leverages character-level subword representations \cite{devlin-etal-2019-bert} or n-grams \cite{bojanowski-etal-2017-enriching} to enable inferring representations for words that are not in the training corpus. The methods developed in this paper differ in that our model learns from text perturbations in the training data and builds word embeddings such that words that are likely perturbations of each other have similar embeddings. We use the FastText model from \citet{bojanowski-etal-2017-enriching} as a baseline and show that our embeddings are more robust to most types of adversarially perturbed text than FastText embeddings, which use the average of a word's character n-grams to encode its meaning.
We choose FastText as a baseline because it can infer embeddings for words not found in its training set, which the original Word2Vec cannot.
Prior work by \citet{piktus-etal-2019-misspelling} also builds on FastText to mitigate misspellings, focusing on training the embeddings to be robust to common character substitutions; our work additionally handles 8 other perturbation types.

\paragraph{Pitfalls of bag-of-characters word representations}
Representing words as bags of characters may seem attractive because it catches misspellings that involve arbitrary character permutations, such as ``advresairal'' as a misspelling of ``adversarial'', as well as character repetition as in ``pleease''. However, bag-of-characters approaches have serious limitations, surpassed in our work. First, in the English dictionary\footnote{defined by Merriam-Webster (\url{https://www.merriam-webster.com}), collected by \url{https://github.com/dwyl/english-words}} of 466,551 words, 312,790 words collide with at least one other word if represented as bags of characters. Each bag of characters maps to 2.03 words on average, and up to 116 words. Second, our method trains embeddings on a specific corpus and can work
with a specific portfolio of perturbations. \textit{A propos}, traditional spam is intentionally misspelled to avoid filters while clickbait titles often replace characters with Unicode look-alikes.

\paragraph{Text perturbation types}
Table~\ref{table:perturbation_types} describes ten types of adversarial text obfuscation---those covered in the literature and those we found in obfuscated public posts on Facebook.
Additional types include adding punctuation or emoji before or after words without spaces (separators), but we delegate them to tokenization and do not include here.

\section{Text Perturbation Defense Methods}

To address each of the ten perturbation types, we introduce three tools: two rule-based defense methods to run directly on input text first, and then a method to build word embeddings robust to the remaining text perturbation types.

\paragraph{Alternating Characters Defense}
We use the first rule-based defense, the Alternating Characters Defense (ACD), to fight the Combined Unicode, Space separation, and Zero-width space separation perturbations.
It first checks for alternating whitespaces and then checks for arbitrary alternating non-alphanumeric characters.
The first step is performed on the entire document and fixes each affected portion by joining adjacent single-character fragments. In the second step, the document is split into words on whitespace, and we check each word for at least half of its characters being non-alphanumeric. In the typical case considered in our empirical studies, all even-numbered or all odd-numbered characters are identical. Because the inserted character needs to appear at least twice to be detected, words of less than three characters are not considered.

\paragraph{Unicode Canonicalization} (UC) is a rule-based defense that counters the Replace Unicode and Tandem character obfuscation perturbations. We define mappings from obfuscated characters to recognizable characters and replace all instances of the obfuscated characters with the correct recognizable characters in the text.
To address both perturbation types, we include mappings for Unicode confusables and for tandem combinations of characters.
In our empirical studies, we use the crowdsourced mappings for Unicode confusables on the Unicode website\footnote{\url{https://www.unicode.org/Public/security/8.0.0/confusables.txt}} and a custom mapping from 38 tandem character combinations to 18 characters\footnote{Examples include /\textbackslash $\rightarrow$ A and () $\rightarrow$ o.}.

\paragraph{Defending Against Split Words}
The rest of this section discusses Continuous Word2Vec (CW2V), our method for building perturbation-robust word embeddings.
CW2V builds embeddings that reflect word spelling and thus relies on a tokenizer to split the document into words, e.g.,
using non-word characters to determine word boundaries (as used in our experiments to ensure reproducibility). So, the Fake Punctuation and Random Spaces perturbations both effectively take a word and split it into multiple words, yielding multiple word embeddings. To evaluate the effectiveness of CW2V against these perturbations, we measuring how close the average of the embeddings of the word-parts is to the embedding of the whole word. Among ideas for future work, we mention combining adjacent words during tokenization to make word boundaries less performance-critical.\footnote{To mitigate the Fake Punctuation perturbation, one can treat punctuation like any other character and not use it to determine word boundaries. The resulting word embeddings should be closer to the de-perturbed word embeddings. However, not using punctuation during tokenization causes more harm to ML performance than do the few words split with these perturbations. Therefore, the experiments in this paper use punctuation in the tokenizer.}

\begin{figure*}
\begin{center}
\includegraphics[width=0.90\textwidth]{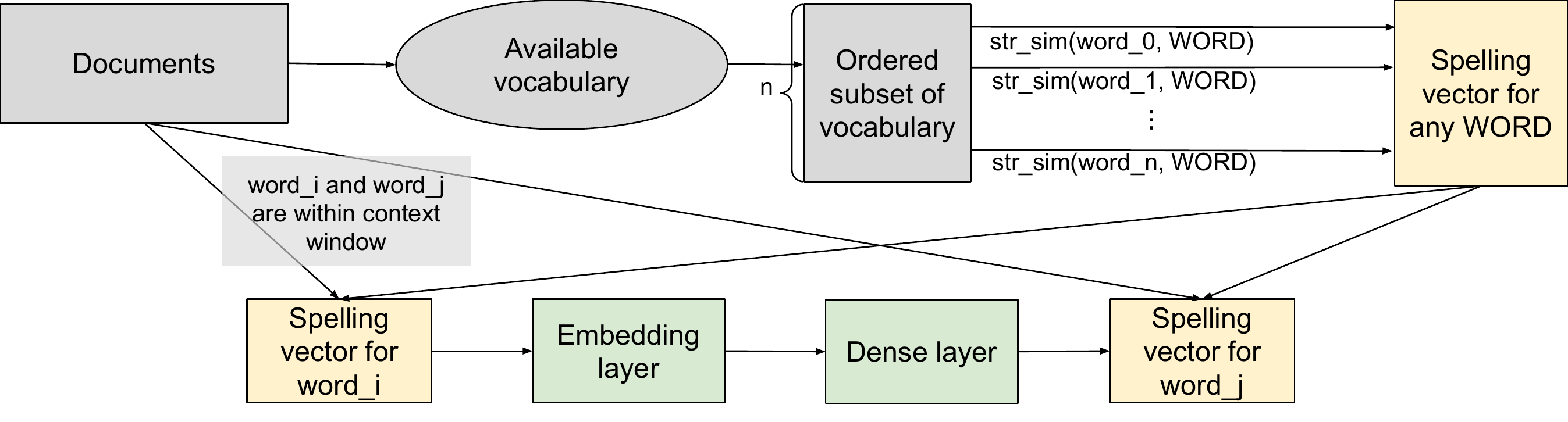}
\end{center}
\vspace{-2mm}
\caption{
\label{figure:method}
Continuous Word2Vec (CW2V). Given a set of training documents, we first build a vocabulary of available words. We then choose a subset of the vocabulary, give it an ordering, and use string similarity to the words in this index as the elements of the spelling vectors. We then use the spelling vectors of words in the training documents that are within a context window of each other as the training input and output of a neural network and learn the parameters in the Embedding layer and Dense layer. We can then infer the embedding of any word by multiplying its spelling vector by the Embedding layer matrix.}
\vspace{-2mm}
\end{figure*}

\paragraph{Word embeddings for continuous words} \hspace{-3mm}
 seek to ensure that small spelling changes result in small changes in the embeddings, in effect, treating words as \textit{continuous} rather than discrete.
Figure~\ref{figure:method} illustrates the proposed method for learning word embeddings.
We first extract a vocabulary from the training corpus and then
select its subset as an index of axes on which we build \textit{spelling vectors} by measuring string similarity (\texttt{str\_sim} below). Then, for words within a context window, we pass the spelling vectors to a shallow neural network to learn word embeddings. We can then infer the embedding of any word by multiplying its spelling vector by the first layer of the neural network.

\paragraph{Selecting an index subset of the vocabulary}
\hspace{-3mm}
is key to ensuring that words with similar spellings have similar spelling vectors. The size of this subset is a hyperparameter, \texttt{n} in Figure~\ref{figure:method}.
We cluster the words in the available vocabulary into \texttt{n} clusters using hierarchical agglomerative clustering, where the distance metric between each pair of words is the Levenshtein edit distance divided by the length of the shorter word.
The index subset of the vocabulary is built by randomly selecting one word from each cluster (alternatively, for each cluster one can find the word that minimizes the sum of distances to other words in the cluster).
This process is similar to the robust encoding method proposed by \citet{roben}. Other than technical differences such as the clustering algorithm and the method for choosing the ``best'' word in each cluster, our method is different in that it is choosing only an index as described here rather than the final word representation; we can then use the index to represent any word outside the training set, and we also take context into account to build the embeddings later on.

\paragraph{Definition of str\_sim}
Given a word $w$, we define its spelling vector as \texttt{str\_sim}$(w_{i}, w)$ where $i$ ranges from 0 to the length of the vocabulary, and $w_i$ is the $i$th element in the ordered subset of the vocabulary.
Given two words as strings $w_a$ and $w_b$, we define
\texttt{str\_sim}$(w_{a}, w_b)$ as
\begin{equation}
\texttt{str\_sim}(w_{a}, w_b) = \begin{cases}
2 * min(len_{a}, len_{b}) & \mbox{if } w_a = w_b \\
\frac{min(len_{a}, len_{b})}{L(w_{a}, w_b)} & \mbox{otherwise } \end{cases} 
\label{eq:str_sim}
\end{equation}
\noindent where $len_a$ and $len_b$ are the lengths of $w_a$ and $w_b$, respectively, and $L(w_{a}, w_b)$ is the Levenshtein edit distance between $w_a$ and $w_b$. This formula defines \texttt{str\_sim} as the reciprocal of Levenshtein distance, scaled to the length of the smaller word (with a special case when the Levenshtein distance is 0). The spelling vector describes a distribution of similarities over the index subset of the vocabulary.

By construction, metric distance between spelling vectors reflects string distance between their original words, and this property extends to further constructs involving these vectors. Thus, we use spelling vectors to train a neural network with one hidden layer. As in the \textit{skip-gram} model by \citet{mikolov2013efficient}, the input to the neural network consists of vectors for $word_i$ and the output consists of vectors for $word_j$, each such word must be within a fixed context window of $word_i$ in the training corpus. In our work (Figure~\ref{figure:method}),  \textit{spelling vectors} of dimension $n$ are used as
the input and output of the neural network. The dimension of the hidden layer is a hyperparameter $h$. Two matrices of sizes $n \times h$ and $h \times n$ are jointly trained using stochastic gradient descent. After training, the final embedding of a word is found by multiplying its spelling vector by the first matrix of the neural network (the Embedding matrix).

\paragraph{Hyperparameters}
Hyperparameters introduced in Figure~\ref{figure:method} include $n$, the size of the index subset of the vocabulary.
Intuitively, a larger $n$ allows for more expressive vectors.
But if we set $n$ to the number of words in a de-perturbed version of the training corpus, then each de-perturbed word may form a cluster, and the spelling vectors would reflect proximity to each word in the de-perturbed vocabulary.
Similarly, if we set $n$ to a fraction of the vocabulary size, then the resulting ordered subset would contain a set of words whose strings are as far apart as possible, which is attractive when computing spelling vectors. For experiments, we
use hyperparameters from the model in \citet{mikolov2013efficient} and \citet{10.5555/2999792.2999959} when learning word embeddings. One is the size of the context window\hush{ within which we collect word pairs as input and output to the neural network}, and other inherited hyperparameters include the size of the hidden layer, learning rate, batch size, and the number of epochs.
The hyperparameter $c$ controls early stopping: if the loss does not decrease for $c$ training epochs in a row, then we stop training. A hyperparameter inherited from \citet{10.5555/2999792.2999959} is $t$, the amount by which we subsample frequent words when selecting them for input into the neural network; to balance the effects of common and uncommon words on the learned parameters, we include each word $w_i$ in the training set with probability $1 - \sqrt{\frac{t}{f(w_i)}}$ where $f(w_i)$ is the frequency of $w_i$ in the training corpus.

\section{Word Vector Validation}

To verify that our embeddings encode both spelling and context, we measure distances between embeddings while varying both.
We measure cosine distances between pairs of embeddings (one minus cosine similarity), which range from 0 to 2; orthogonal embeddings have distance 1.

For these results, we train embeddings on a set of 40K randomly selected, publicly available, and de-identified Facebook posts, passed through the Alternating Characters Defense and Unicode Canonicalization.
The posts are lowercased and tokenized, after which punctuation and emoji are dropped. We compare with FastText embeddings trained on the same preprocessed corpus with the same hyperparameters; the hidden sizes and embedding dimensions for both methods are 200.
The spelling vectors for our embeddings are 0.005 of the vocabulary size, about 385.

\begin{table}[tb]
    \centering
    \begin{tabular}{|l|ll|}
        \hline
         & Facebook posts & English dictionary \\
         \hline
        CW2V & \textbf{0.718} & \textbf{0.772} \\
        FastText & 0.005 & 0.046 \\
        \hline
    \end{tabular}
    \caption{
    \label{tab:spelling_dists}
    Correlation between Levenshtein and cosine distances for FastText and our method CW2V. Embeddings are trained on a corpus of Facebook posts.
    Results are based on 100 words from Facebook posts and 100 dictionary words.}
\end{table}

\paragraph{Correlation with spelling}
To confirm that words with similar spelling have similar embeddings, Table~\ref{tab:spelling_dists} shows the correlation between word pairs' Levenshtein distances and their embeddings' cosine distances.
We compare results between CW2V and prior high-quality embeddings from FastText. 
For CW2V, for a randomly selected set of 100 words from the English dictionary, this correlation is 0.772, and for a randomly selected set of 100 words from the Facebook post training corpus, this correlation is 0.718. These correlations are much
higher than respective numbers for FastText embeddings trained on the same corpus --- 0.005 and 0.046.

\begin{table*}[ht]
\begin{center}
\begin{tabular}{|l|llll|}
\hline
 & Neighboring key & Transposition & Vowel rep \& del & Random spaces \\
\hline
CW2V Facebook posts & \textbf{0.015} & \textbf{0.175} & \textbf{0.045} & 0.502 \\
FastText Facebook posts & 0.230 & 0.351 & 0.172 & \textbf{0.320} \\
\hline
CW2V English dictionary & \textbf{0.110} & \textbf{0.043} & 0.217 & 1.036 \\
FastText English dictionary & 0.224 & 0.320 & \textbf{0.140} & \textbf{0.391} \\
\hline
\end{tabular}
\caption{
\label{table:perturb_dists}
Average cosine distances \hush{(1 - cosine similarity)} between embeddings of words and their perturbed versions, divided by average distances between embeddings of random words. Smaller numbers indicate that embeddings are closer for perturbations than random words; the scores where perturbations make less difference are in bold. CW2V (our method) and FastText embeddings are trained on Facebook posts. We show results on words from the English dictionary and from Facebook posts.}
\end{center}
\vspace{-1mm}
\end{table*}

\paragraph{Effects of text perturbation on embeddings}
In addition to comparing embeddings of dictionary words with similar spelling, a major motivation for CW2V was to ensure
that word perturbations do not alter embedding vectors by large amounts. Table~\ref{table:perturb_dists} shows the average distances between the embeddings of words and their perturbed versions, divided by the average distance between embeddings of random words; a smaller number indicates that the perturbation  embeddings are closer for the perturbation than random words, and 1 would indicate that the embeddings are just as different for perturbations as for random words.
Distances are measured using 500 randomly selected dictionary words and 500 randomly selected words from the training corpus, as indicated. We show average distances between an original word and the word perturbed by the Neighboring Key perturbation, the Transposition perturbation, the Vowel Repetition and Deletion perturbation, and the Random Spaces perturbation. For Random Spaces, we randomly add spaces, and then average the embeddings for the resulting ``words''; the Fake Punctuation perturbation will likely show the same effect.

The scores in Table~\ref{table:perturb_dists} show that for words perturbed by the Neighboring Key and Transposition perturbations, the embeddings are closer to the original word embeddings with CW2V than with FastText.
For the Vowel Repetition and Deletion perturbation, the embeddings are closer when the embeddings are trained and tested on the same dictionary. For the Random Spaces perturbation (and therefore also Fake Punctuation perturbation), the FastText-trained embeddings are closer than the CW2V embeddings, so FastText may be the preferred method when these two perturbations are the majority.
Because the Neighboring key, Transposition, and Vowel Repetition and Deletion perturbations are more common organically than the Random Spaces perturbation, and because typical use cases for these embeddings involve training and testing on the same dictionary, we conclude that the embedding computation proposed in this paper is more robust to text perturbation.
%
%
\section{Engagement Bait Classifier Performance}

\begin{figure}[bt]
\includegraphics[width=0.47\textwidth]{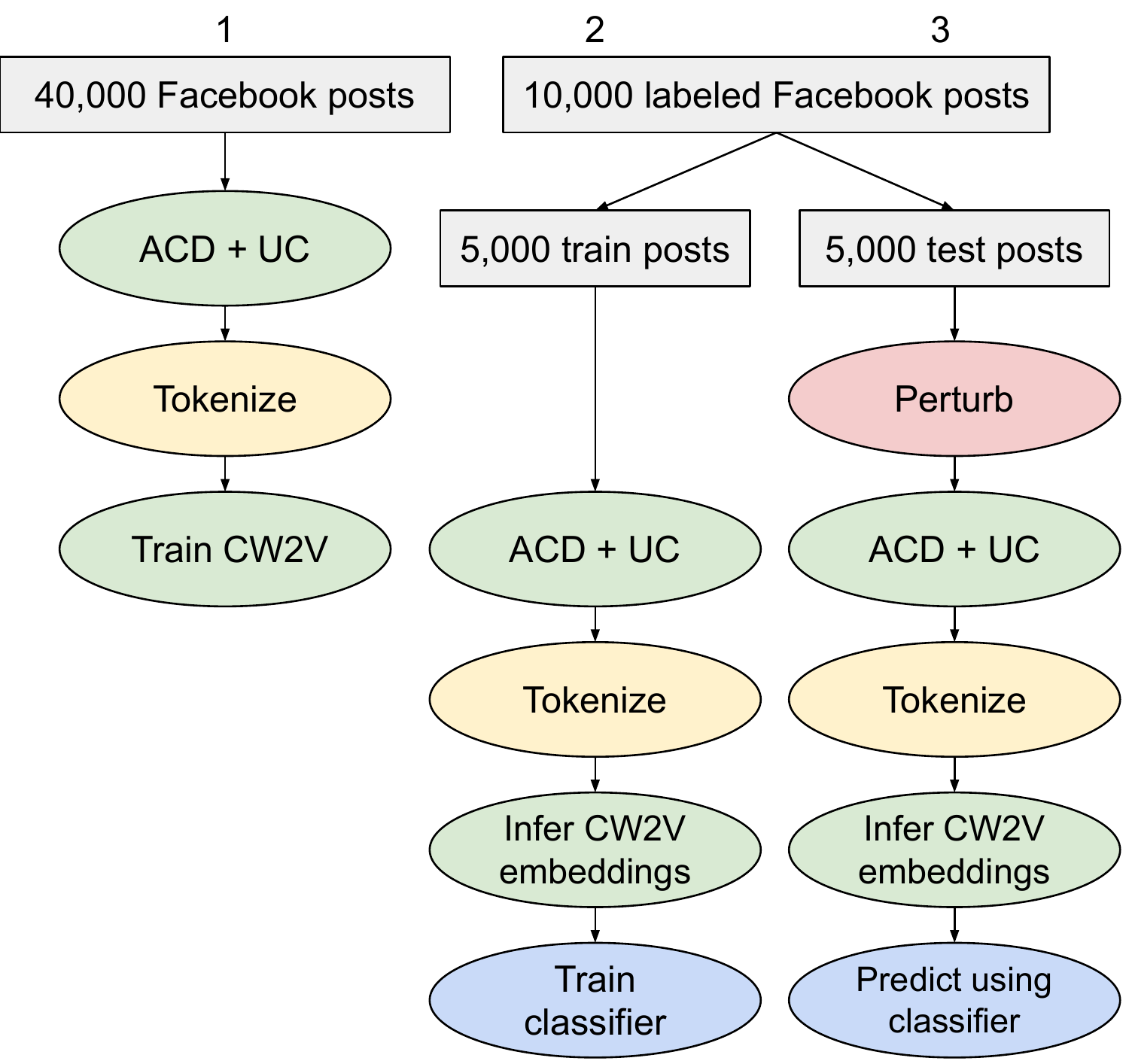}
\caption{\label{figure:pipeline}
Proposed deobfuscation and classification pipelines. 40K randomly selected, publicly available, and de-identified Facebook posts are passed through rule-based defenses (ACD and UC), then used to train word embeddings according to CW2V, see Figure~\ref{figure:method}. Corresponding pipelines for classifier training and prediction are described.
\hush{
We then select 10K labeled Facebook posts and split them for training and evaluation. If the training set is passed through the two rule-based string defenses, then that is done before the training text is tokenized, its embeddings are inferred, and the downstream classifier is trained. If the test set is perturbed, then that is done first, and if it is passed through the two rule-based string defenses, then that is done next. Then the test set is tokenized, its embeddings are inferred, and it is passed to the downstream classifier for prediction.
}
}
\end{figure}

\begin{table*}[ht]
\begin{center}
\begin{tabular}{|l|ll|}
\hline
 & Original & Perturbed \\
 & AUC & AUC \\
\hline
ACD, UC, and CW2V & $0.704 \pm 0.011$ & $0.673 \pm 0.012$ \\
ACD, UC, and BERT & $0.714 \pm 0.015$ & $0.649 \pm 0.016$ \\
BERT & $0.713 \pm 0.014$ & $0.612 \pm 0.009$ \\
ACD, UC, and FastText (5K) & $0.763 \pm 0.003$ & $0.723 \pm 0.003$ \\
FastText (5K) & $0.761 \pm 0.003$ & $0.632 \pm 0.003$ \\
ACD, UC, and FastText (10K) & $0.792 \pm 0.002$ & $0.751 \pm 0.003$ \\
FastText (10K) & $0.790 \pm 0.002$ & $0.635 \pm 0.003$ \\
ACD, UC, and Adv.Tr. FastText & $0.787 \pm 0.002$ & $0.750 \pm 0.003$ \\
Adv.Tr. FastText & $0.786 \pm 0.002$ & $0.649 \pm 0.004$ \\
\hline
\end{tabular}
\caption{
\label{table:pipeline_results}
Effects of perturbations and defenses on downstream classifiers. Original AUC and Perturbed AUC are the areas under the ROC curve for the original test set and perturbed test set, respectively. ACD and UC indicate that the training and test sets were passed through our respective rule-based defense mechanisms. CW2V indicates classification based on our CW2V embeddings. Adv.Tr. FastText is a FastText model adversarially trained with 5K original posts and 5K perturbed posts.}
\end{center}
\end{table*}

How well our embeddings encode word context is evaluated through the accuracy of downstream classifiers.
We propose a classification pipeline that addresses all perturbations classified in Table \ref{table:perturbation_types}.
First, it performs the two defenses: Alternating Characters Defense and Unicode Canonicalization.
After tokenizing the text, it computes the continuous word embeddings for individual words, and uses them as features to train a classifier. We test our pipelines with and without artificial perturbation and our two defenses, and compare the performance with state-of-the-art methods.

\paragraph{Engagement Bait Classifier}
For evaluation, we choose a downstream ML model based on logistic regression that classifies Facebook posts as \textit{engagement bait} or not. Engagement bait is content asking for \textit{like}s, \textit{comment}s, etc. We train CW2V embeddings on a random sample of 40K publicly available Facebook posts in English.
The classifier is trained and tested on a random de-identified sample of publicly available Facebook posts.
The train and test sets for the classifier are manually annotated for engagement bait.\footnote{Each post was independently annotated by two annotators. If they disagreed, a third annotation was used for a majority vote.} For this classifier, the features were only based on the text as described here; most industry production classifiers likely use other features such as engagement metrics, images, and user-based features to increase accuracy.

\paragraph{Classification Pipelines Tested}
Table~\ref{table:pipeline_results} shows the area under the ROC curve for the downstream classifier with and without defenses and added perturbations.
Our proposed pipeline includes ACD and UC, and we use the elementwise average of the CW2V word embeddings in each post as features for the downstream classifier.
For comparison to state-of-the-art models, we finetune a BERT (base, uncased) classifier \cite{devlin-etal-2019-bert} on the training set, and train a classifier that uses the elementwise average of FastText embeddings as features.
We train FastText embeddings on 5K and 10K posts.
For each result, the embeddings are trained 10 times, and each set of trained embeddings is used to train and test 10 classifier models, resulting in 100 runs.
For comparison to adversarial training, we train FastText embeddings on a set of 5K perturbed and 5K original posts, also with 100 runs over 10 FastText embedding sets.
For a fair comparison, all CW2V and FastText embeddings have 200 elements.

Word embeddings are trained on organic unperturbed Facebook posts in English, passed through the ACD and UC defenses when indicated (with the exception of the adversarially trained FastText embeddings, which are trained on 5K perturbed posts and 5K unperturbed posts).
\hush{For the results without perturbation, we assume that the organic posts are not perturbed (true for the majority of posts) and perform training and testing as usual.}
Using a perturbation selected randomly from Table~\ref{table:perturbation_types}, we perturb each word (longer than two characters) in the test set.
When ACD and UC are applied, they are applied to both the train set and test set.
When perturbations and defenses are both applied, the perturbations happen before the defenses.

\hush{The first row of Table~\ref{table:pipeline_results} named ``ACD, UC, and CW2V'' represents our pipeline; we pass all the training and testing data through ACD and UC, and use our proposed embeddings for classification.
The row labeled ``ACD, UC, and FastText'' shows a similar pipeline, with ACD and UC, but the word embeddings are calculated using FastText.
We see that the pipeline performs the best with FastText, since it learns the most parameters overall, leading to a relative increase in information.
However, perturbed input decreases the score more with FastText than with CW2V.
The row labeled ``FastText'' simply uses FastText embeddings for classification, without ACD or UC.
We can see that perturbation decreases the score much more with this method, showing the effectiveness of ACD and UC.}

We see that applying ACD and UC helps mitigate dips caused by perturbation in the test set.
CW2V is less affected by perturbation than BERT. However, FastText performs the best overall, likely because it learns more parameters overall, leading to a relative increase in information. Both methods produce 200-dimensional embeddings, but FastText stores an embedding for each character ngram while training (resulting in $200 * a^n$ parameters, for $a$ distinct characters and character ngrams up to $n$ characters), while CW2V stores no embeddings and calculates the test set word embeddings using a learned matrix (which has $200 * len(spelling embedding)$ parameters).

\begin{figure}[tb]
\includegraphics[width=0.4\textwidth]{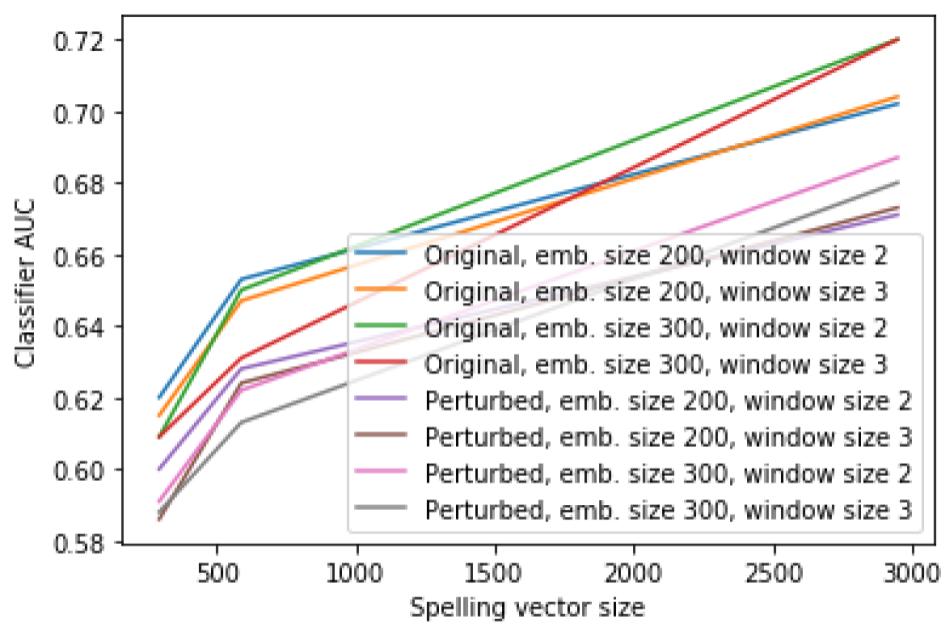}
\caption{\label{figure:performances}
Classifier performance with varying hyperparameters, based on spelling vector size.}
\vspace{-2mm}
\end{figure}

\paragraph{Window Size and Embedding Size}
Figure~\ref{figure:performances} shows the performance of the pipeline with ACD, UC, and CW2V for the original pipeline (1) as is and (2) with additionally perturbed data (defended through the pipeline). It compares the AUC of the classifier with 200- and 300-dimensional embeddings, and varies the window size when training embeddings between 2 and 3 words. We see that the window size and embedding dimensionality do not affect the score nearly as much as the size of the spelling vector.

\section{Future Work}

We discuss modifying our technique to increase accuracy with respective tradeoffs in computational complexity, loss of information, and implementation difficulty.

First, we can add 26 ``alphabet words'' to the index subset of the vocabulary for spelling vectors: ``aaaa'', ``bbbb'', ... ``zzzz''. This would distinguish spelling vectors at a finer-grained level, i.e., without these 26 anchors, the words ``probable'' and ``provable'' will likely 
map to the same spelling vector. We could extend the idea by adding bigrams most frequent in the text, such as ``efefef'' and ``ghghgh''. The tradeoff would be increased computational complexity.

Second, we could modify the distance metric for spelling-index clustering to incorporate cosine distances between FastText vectors to allow a tradeoff between variance in the resulting vectors based on spelling and based on meaning. However, we showed that perturbation affects distances between FastText vectors in an undesirable way, so we would need to limit the spelling index clusters to only dictionary words, a considerable tradeoff.

Third, instead of \texttt{str\_sim}, we could use one minus the Jaccard distance to determine spelling vectors. However, this does not account for the order of letters and therefore maps anagrams to the same vectors; we could use a linear combination of Jaccard distance and \texttt{str\_sim} in order to give more weight to order-independent metrics.

Fourth, as mentioned earlier, we could use cluster centers when selecting the spelling index, rather than a random element of each cluster. To implement this improvement, we need an efficient method to find, within each cluster, the word with the smallest total distance to other words.

Other future work includes combining our methods with language models by changing the language model's features from discrete token IDs to continuous-valued vectors.

\section{Conclusion}
Classifying posts with adversarial text perturbation remains a challenge. Rather than continue the line of work ``fixing'' misspelled words in user input text \cite{zhou-etal-2019-learning,10.5120/ijca2019919384,fivez-etal-2017-unsupervised}, we take inspiration from methods that modify context-based word embeddings to address character substitution \cite{piktus-etal-2019-misspelling} and build word embeddings that reflect both the meanings and spellings of words.

We classified adversarial text perturbation into ten common types and proposed mitigation strategies.
We showed the effectiveness of two rule-based string manipulations in defending against five of the perturbations.
We also proposed a method of calculating word embeddings and measured the embeddings' robustness to the other five perturbations.
We demonstrated that a downstream classifier is less vulnerable to adversarial text perturbation when we use these mitigation strategies.

Our work fits in with related efforts in the industry, which include measuring the prevalence of adversarial text perturbations, measuring the effects of perturbation tools and defenses, and improving model interpretability \cite{captum2019github}.
The proposed defenses can be used in settings where adversaries try to avoid detection through text perturbation, or when user-written text is susceptible to misspellings.
Our proposed classification pipeline is a step in the direction of robust industry solutions.
\section*{Acknowledgements}
We gratefully acknowledge Damon McCoy for substantial guidance and feedback.
\bibliography{ms}
\end{document}